\documentclass[letterpaper, 10 pt, conference]{ieeeconf}  % Comment this line out if you need a4paper

\IEEEoverridecommandlockouts                              % This command is only needed if 
\usepackage[T1]{fontenc}
\usepackage[caption=false, font=footnotesize]{subfig}

\usepackage{graphics}           
\usepackage{times}              
\usepackage{amsmath}            
\usepackage{amssymb}            
\usepackage{graphicx}
\usepackage{algorithm}
\usepackage[noend]{algpseudocode}
\usepackage{booktabs}
\usepackage{color}
\usepackage{multirow}
\usepackage{setspace}
\definecolor{instructioncolor}{rgb}{.5,.5,.5}
\usepackage{threeparttable}

\makeatletter
\renewcommand{\maketag@@@}[1]{\hbox{\m@th\normalsize\normalfont#1}}%
\makeatother

\usepackage[font=small]{caption}

\def\secref#1{Sec.~\ref{#1}}
\def\figref#1{Fig.~\ref{#1}}
\def\tabref#1{Tab.~\ref{#1}}
\def\eqref#1{Eq.~(\ref{#1})}
\def\algref#1{Alg.~\ref{#1}}

\makeatletter
\usepackage{xspace}
\DeclareRobustCommand\onedot{\futurelet\@let@token\@onedot}
\def\@onedot{\ifx\@let@token.\else.\null\fi\xspace}

\def\etal{{et al}\onedot}
\makeatother

\def\etalcite#1{\etal~\cite{#1}}

\usepackage{array}
\newcolumntype{L}[1]{>{\raggedright\let\newline\\\arraybackslash\hspace{0pt}}m{#1}}
\newcolumntype{C}[1]{>{\centering\let\newline\\\arraybackslash\hspace{0pt}}m{#1}}
\newcolumntype{R}[1]{>{\raggedleft\let\newline\\\arraybackslash\hspace{0pt}}m{#1}}

\renewcommand{\b}[1]{\mbox{\boldmath$#1$}}

\renewcommand{\v}[1]{{\b #1}} 

\usepackage{cite}
\usepackage{graphicx} % for pdf, bitmapped graphics files
\usepackage{epstopdf}
\usepackage{epsfig} % for postscript graphics files
\usepackage{amsmath} % assumes amsmath package installed
\usepackage{amssymb}  % assumes amsmath package installed

\usepackage{algorithm}
\usepackage{algorithmicx}
\usepackage{algpseudocode}
\usepackage{multirow}
\usepackage{xcolor}
\usepackage{url}

\title{\LARGE \bf
	ElC-OIS: Ellipsoidal Clustering for Open-World Instance Segmentation on LiDAR Data
}

\author{Wenbang Deng \and Kaihong Huang \and Qinghua Yu \and Huimin Lu$^*$ \and Zhiqiang Zheng \and Xieyuanli Chen$^*$% <-this % stops a space
  \thanks{All authors are with the College of Intelligence Science and Technology, National University of Defense Technology, Changsha, China.}%
  \thanks{$^*$corresponding authors, \{lhmnew, xieyuanli.chen\}@nudt.edu.cn}
\thanks{This work was supported in part by the National Science Foundation of China under Grant U1913202, and U22A200600, as well as Major Project of Natural Science Foundation of Hunan Province under Grant 2021JC0004.
}%
}

\graphicspath{{figs/}}

\begin{document}
	
	\maketitle
	\thispagestyle{empty}
	\pagestyle{empty}
	
	%%%%%%%%%%%%%%%%%%%%%%%%%%%%%%%%%%%%%%%%%%%%%%%%%%%%%%%%%%%%%%%%%%%%%%%%%%%%%%%%
	\begin{abstract}
	Open-world Instance Segmentation~(OIS) is a challenging task that aims to accurately segment every object instance appearing in the current observation, regardless of whether these instances have been labeled in the training set. This is important for safety-critical applications such as robust autonomous navigation. In this paper, we present a flexible and effective OIS framework for LiDAR point cloud that can accurately segment both \textit{known} and \textit{unknown} instances (i.e., seen and unseen instance categories during training). It first identifies points belonging to known classes and removes the background by leveraging close-set panoptic segmentation networks. Then, we propose a novel ellipsoidal clustering method that is more adapted to the characteristic of LiDAR scans and allows precise segmentation of unknown instances. Furthermore, a diffuse searching method is proposed to handle the common over-segmentation problem presented in the known instances. With the combination of these techniques, we are able to achieve accurate segmentation for both known and unknown instances.
	We evaluated our method on the SemanticKITTI open-world LiDAR instance segmentation dataset. The experimental results suggest that it outperforms current state-of-the-art methods, especially with a 10.0\% improvement in association quality. The source code of our method will be publicly available at \url{https://github.com/nubot-nudt/ElC-OIS}.
	\end{abstract}

	%%%%%%%%%%%%%%%%%%%%%%%%%%%%%%%%%%%%%%%%%%%%%%%%%%%%%%%%%%%%%%%%%%%%%%%%%%%%%%%%
	\section{Introduction}
	Precise point-cloud instance segmentation is crucial for autonomous cars and robots to accurately understand their surrounding environments. It provides valuable information for essential tasks such as obstacle avoidance, path planning, and navigation. Recently,  LiDAR-based semantic segmentation~\cite{milioto2019iros,li2022ral} and panoptic segmentation~\cite{hong2021cvpr, sirohi2022tro} have been widely studied.
	These methods aim to predict the semantic category for each point and/or cluster instances in LiDAR data. However, they only focus on the \textit{known} instances, i.e., objects that are annotated in the training data, such as people and vehicles. Typically, they are unable to identify other \textit{unknown} instances, i.e., unlabeled during training, yet frequently appear in real-world scenarios. This incapacity to recognize long-tail unknown objects for autonomous mobile systems in open worlds can lead to serious crises. For example, if the vehicle fails to recognize a baby stroller that is not labeled in the training data, it may not be able to brake or avoid a collision, leading to severe consequences.
	
	\begin{figure}[t]
		\centering
		\includegraphics[width=8.5cm]{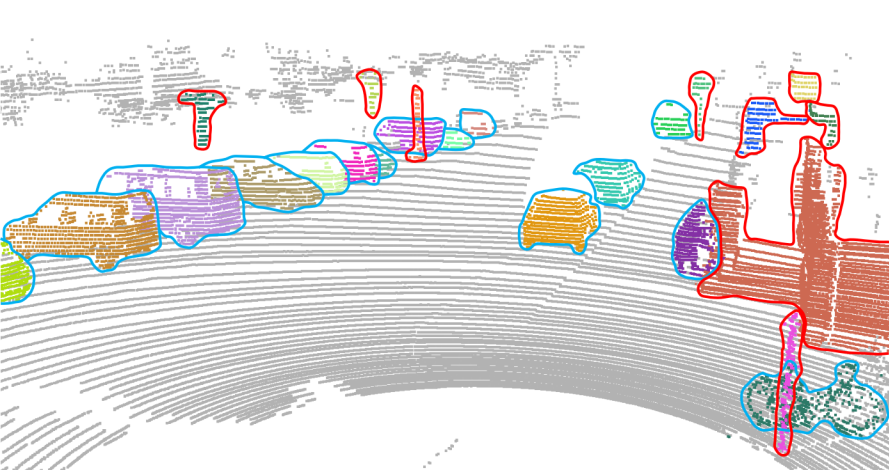}
		\caption{Visualization of our open-world instance segmentation results. Different instances are marked with unique colors. The instances circled with red lines are unknown instances, i.e., instances unlabeled in the training set. The left ones circled with blue lines are known instances, such as car, person, and bicycle.}
		\label{fig:complete-result}
	\end{figure}
	
	Tackling the open-world instance segmentation (OIS) problem is highly challenging, as traditional end-to-end deep learning-based methods fail in this task by definition. Only a few methods exist that either use the "objectness" learned from known instance segmentation networks to choose unknown instances from candidates~\cite{hu2020ral}, or generate instance proposals with traditional clustering methods and then use unsupervised learning to remove the background~\cite{nunes2022ral}. The former has relatively weak generalization ability for unknown instance segmentation, as the learned "objectness" may over-fit to known instances used for training. Meanwhile, the latter does not utilize existing known instance information provided in the training set. Both of these methods cannot accurately achieve LiDAR OIS.
	
	In this paper, we propose a novel, flexible, and effective LiDAR OIS framework that fully utilizes available information to accurately segment both known and unknown instances, as shown in~\figref{fig:complete-result}. The framework consists of three main components: i) close-set panoptic segmentation, ii) unknown instance clustering, and iii) known instance refinement. It first leverages existing information by using a close-set panoptic segmentation network to generate raw point-wise semantic labels and known object instances. The second component proposes a novel ellipsoidal clustering method to segment unknown instances effectively where points do not belong to a known instance or background. Finally, we propose a diffuse searching method for refining the over-segmented instances.
	We thoroughly evaluate our method on the SemanticKITTI open-world LiDAR instance segmentation benchmark~\cite{nunes2022ral} and compare it with other state-of-the-art methods.
	The experimental results validate the effectiveness of our proposed LiDAR OIS framework and show the superiority compared to the state-of-the-art methods.

	To sum up, the contributions of our paper are threefold:
	\begin{itemize}
		\item We propose a novel LiDAR OIS method that outperforms the state-of-the-art methods by 10\% and achieves the best result on SemanticKITTI LiDAR OIS benchmark.
		\item We propose a novel ellipsoidal clustering method that outperforms other baseline methods in class-agnostic instances clustering on LiDAR data.
		\item Our OIS framework is highly modular and flexible, which works well with different panoptic segmentation and/or clustering methods.
	\end{itemize}

	\begin{figure*}[th]
		\centering
		\includegraphics[width=1\textwidth]{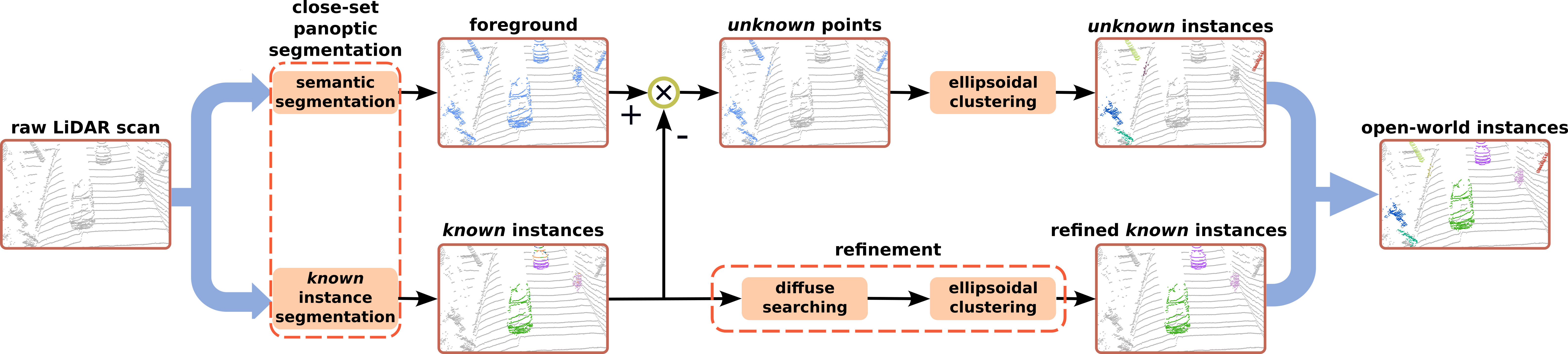}	
		\caption{Overview of our framework. We first obtain the semantic information from the semantic segmentation part and the raw known instances from the known instance segmentation part. The semantic information is then used to remove the background points. Next, we remove the raw known instances from the left foreground points, leaving the rest for unknown instances clustering. The rest points are clustered via the proposed ellipsoidal clustering method to generate unknown instances. The known instances are refined by the proposed diffuse searching method. Finally, we combine the refined known instances and unknown instances to generate complete OIS results.}
		\label{fig:overview}
	\end{figure*}

	\section{Related Work}	
	While there have been several works addressing OIS in the computer vision domain~\cite{wang2022cvpr, hwang2021cvpr, wang2021cvpr, wang2021iccv}, researches in this area for LiDAR point clouds are rare.
	Most existing works in LiDAR segmentation focus on semantic~\cite{milioto2019iros}, instance~\cite{zhang2020icra, chen2022tiv}, and combined panoptic~\cite{marcuzzi2023ral, aygun2021} levels. 
	Whereas these methods only focus on handling several classes of instances labeled in the training set. Consequently, they do not account for the presence of important unlabeled instances in the real world, making them unable to achieve OIS.
	
	Class-agnostic point-cloud clustering can be seen as a preliminary step for OIS that groups together points belonging to the same object instance, regardless of their classes.
	One well-known clustering solution is to iteratively merge points within a fixed radius in the Euclidean space~\cite{rusu2011icra, klasing2008icra}.
	These methods ignore the characteristic of laser scans, where the farther the point from the sensor, the larger the gap between neighboring points, thus leading to sub-optimal clustering results.
	Douillard~\etalcite{douillard2011icra} utilize voxel grids to represent dense point clouds, where the means, variances, and the density of each voxel grid are regarded as the clustering criterion. While it outperforms Euclidean-based methods, the clustering precision is impacted by voxel resolution. Therefore, the performance will degrade when decreasing the resolution to meet online requirements.
	To improve the execution speed, several works~\cite{bogoslavskyi2016iros, zermas2017icra, chen2019iros} project point clouds into the range images or occupied grid cells~\cite{paigwar2020iros, behley2013iros} to cluster the neighboring pixels/grids meeting certain conditions in 2D space.
	However, such projection-based methods suffer from data loss.
	\cite{che2018jprs, huang2019rs, klasing2009icra, jagannathan2007pami} utilize region growing methods to cluster adjacent points with similar geometrical features such as normal and curvatures.
	The effectiveness of these methods relies on the accuracy of normal and curvature estimation, and the calculation is time-consuming.
	Some works consider the density of point clouds as the clustering criterion~\cite{ester1996kdd, campello2013pakdd}, where the density of an object should be large enough. These methods also require multiple distance calculations.
	Curved-voxel clustering method~\cite{park2019iros} allocates points into different curved voxels according to their spherical coordinates and clusters points in nearby voxels.
	The scope of each curved voxel is determined by the unit size parameters for each spherical direction and dynamically adjusted in proportion to the distance from the sensor.
	However, the curved voxels are segmented roughly to cover the whole spherical space, while bringing in extra points at the corner of the curved voxels.
	Besides, the neighbors of each point in one voxel are the same and lack of flexibility.
	
	Deep neural networks have demonstrated great potential in LiDAR semantic recognition~\cite{nunes2022seg,chen2021mos,jin2022iros}, but their end-to-end learning approach poses a challenge when attempting to identify unknown objects that were not present during training. As such, combining neural networks with traditional clustering techniques represents a promising direction for OIS.
	Wong~\etalcite{wong2020corl} first propose a category-agnostic instance embedding network to encode points into an embedding space. This network also predicts the prototypes of each predicted known things anchor and known stuff, while the points that are not associated are considered unknown points.
	These points are then clustered by DBSCAN~\cite{ester1996kdd} according to the distances in the embedding space and the Euclidean space.
	Nunes~\etalcite{nunes2022ral} first utilize HDBSCAN~\cite{campello2013pakdd} to cluster raw instances from the ground-removed point clouds and leverage self-supervised representation learning to extract point-cloud features.
	A graph is built according to the feature affinity, and this method finally applies a min-cut to the graph to distinguish the foreground and background points.
	Hu~\etalcite{hu2020ral} iteratively cluster instances with different distance thresholds to make up the hierarchical groups of candidate instances.
	The "objectness" of each candidate instance is then estimated by using the PointNet++~\cite{qi2017nips} model, and the candidates with the highest "objectness" scores are considered the final instances.
	Chen~\etalcite{chen2021ral} use the color information and the geometric features of point clouds to train a class-agnostic network for region growing segmentation.
	While these methods show good potentials, they are unable to fully leverage the information provided by segmentation networks and produce accurate clustering results to achieve OIS on LiDAR data.
	
	Different from the abovementioned methods, our proposed method is a flexible and effective open-world LiDAR instance segmentation framework. It first removes the background by exploiting the panoptic segmentation results from the network, then uses a novel ellipsoidal neighbor to better cluster LiDAR scans and realize accurate OIS.
	
	\section{Our Approach}
	We propose a flexible and effective open-world instance segmentation framework, together with novel ellipsoidal clustering and diffuse searching methods.
	\secref{sec:framework} provides an overview of our proposed LiDAR OIS framework.
	\secref{sec:ellipsoidal-neighbor} presents the proposed ellipsoidal neighbor searching, and \secref{sec:ellipsoidal-clustering} details the corresponding clustering method.
	\secref{sec:refinement} introduces the diffuse searching method used to refine the over-segmented known instances.
	
	\subsection{Open-World Instance Segmentation Framework Overview}
	\label{sec:framework}
	
	The overview of our framework is shown in~\figref{fig:overview}. It takes the raw LiDAR scan as input and applies a close-set panoptic segmentation on the LiDAR scan, achieving both semantic and known instance segmentation. We use the semantic information to exclude certain background classes that do not contain instances, such as road, parking, and sidewalk. Then, we subtract the known instance points provided by the panoptic segmentation from the remaining foreground points, leaving only the points that may belong to unknown instances. After that, we utilize the proposed ellipsoidal neighbor-based clustering method to generate unknown instances.
	Scattered and over-segmented instances often result from existing close-set panoptic segmentation. To address this issue, we introduce the diffuse searching method to gather scattered clusters for further refinement with our ellipsoidal clustering method. In the end, we integrate the unknown instances with the refined known instances to obtain a comprehensive OIS result.
	
	Our framework is highly modular, allowing for flexible adjustments such as changing the close-set panoptic segmentation and/or clustering methods. Additionally, it also works without using panoptic segmentation, as our clustering method is class-agnostic and can cluster both known and unknown instances. We have thoroughly evaluated our framework by testing different panoptic segmentation and clustering methods detailed in~\secref{sec:as}. Here, we test two off-the-shelf panoptic segmentation methods, DS-Net~\cite{hong2021cvpr} and Panoptic-PolarNet~\cite{zhou2021cvpr}. They both provide semantic and known instance segmentation results. For more details about panoptic segmentation, we refer to the original papers~\cite{zhou2021cvpr, hong2021cvpr}.
	
	\begin{figure}[t]
		\centering
		\subfloat[]{
			\includegraphics[width=4.8cm]{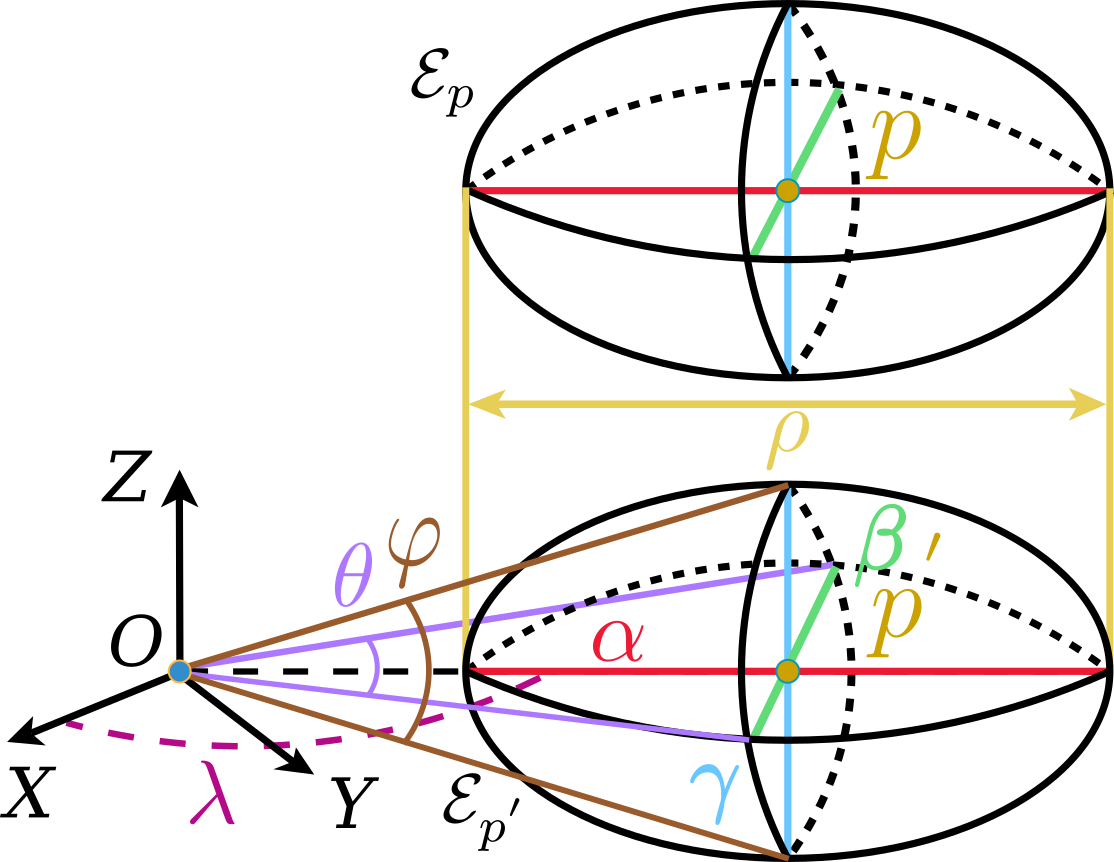}
			\label{fig:ellipsoidal-perspective}
		}%
		\subfloat[]{
			\includegraphics[width=3cm]{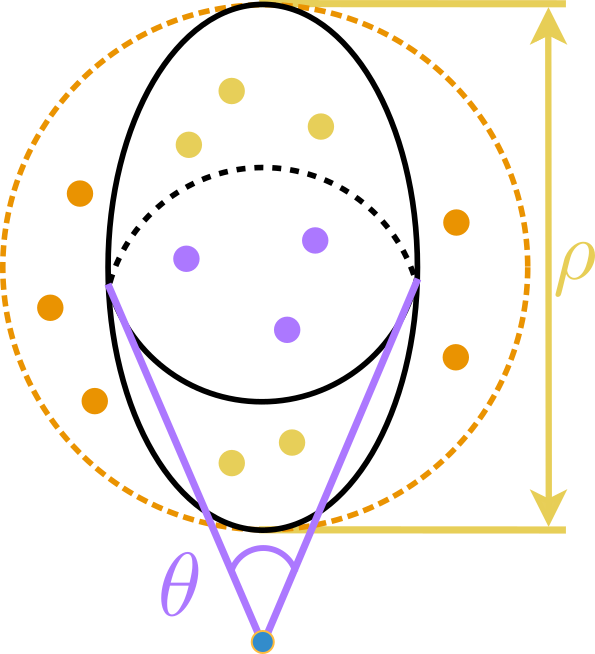}
			\label{fig:ellipsoidal-bev}
		}%
		\caption{The ellipsoidal neighbor searching. (a) The perspective view. (b) The bird's eye view. We assume that $\alpha$ and $\beta$ are the longest and the shortest ellipsoidal axes in this situation. The orange circle indicates the searching range of Kd-tree.}
		\label{fig:ellipsoidal}
		\vspace{-0.2cm}
	\end{figure}
	
	\begin{algorithm}[t]
		\caption{Ellipsoidal Clustering}
		\label{alg:ellipsoidal-clustering}
		\begin{algorithmic}[1]
			\Require Point cloud $\mathcal{P}$, ellipsoidal parameter $\rho$, $\theta$, and $\varphi$
			\Ensure List of instance ID $\mathcal{L}_{ID}$
			
			\State \textbf{Initialization:} $\mathcal{L}_{ID} \leftarrow \{0\}_{\mathcal{P}\text{.size()}}$, list of extended flags $\mathcal{L}_{EF} \leftarrow \{\text{False}\}_{\mathcal{P}\text{.size()}}$, current instance ID $N = 1$, Kd-tree $T_{\mathcal{P}}$ for $\mathcal{P}$, graph of Instance ID $G$, list of indexes of points needed to be queried $\mathcal{L}_Q$
			\For {each point $\v{p}_{i_p} \in \mathcal{P}$}
				\If {$\mathcal{L}_{EF}[i_p] = \text{True}$}
					\State \textbf{continue}
				\EndIf
				\State $\mathcal{L}_Q\text{.push}(i_p)$
				\State \textbf{AddGraphEdge}($N$, $N$, $G$)
				\While {$\mathcal{L}_Q \neq \emptyset$}
					\State $i_q$ = $\mathcal{L}_Q\text{.pop()}$
					\If {$\mathcal{L}_{EF}[i_q] = \text{True}$}
					\State \textbf{continue}
					\EndIf
					\State $a$, $b$, $c$ = \textbf{GetEllipsoidalAxes}$(\v{p}_{i_q}, \rho, \theta, \varphi)$
					\State $\mathcal{L}_{Q_n} = T_{\mathcal{P}}.\textbf{FindNeighbor}(\v{p}_{i_q}, \text{max}(a, b, c))$
					\For {each index $i_n \in \mathcal{L}_{Q_n}$}
						\If {$\v{p}_{i_n}$ $\notin$ ellipsoidal neighbor $\mathcal{E}_{\v{p}_{i_q}}$}
							\State \textbf{continue}
						\EndIf
						\If {$\mathcal{L}_{ID}[i_n] \neq 0$ and $\mathcal{L}_{ID}[i_n] \neq \mathcal{L}_{ID}[i_q]$}
							\State $\textbf{AddGraphEdge}(\mathcal{L}_{ID}[i_n], \mathcal{L}_{ID}[i_q], G)$
						\EndIf
						\If {$\mathcal{L}_{EF}[i_n] = \text{True}$}
							\State \textbf{continue}
						\EndIf
						\State $\mathcal{L}_{ID}[i_n] = N$
						\If {$\textbf{Distance}(\v{p}_{i_q}, \v{p}_{i_n}) \leq \text{min}(a, b, c)$}
							\State $\mathcal{L}_{EF}[i_n] = \text{True}$
						\Else
							\State $\mathcal{L}_Q\text{.push}(i_n)$
						\EndIf
					\EndFor
				\EndWhile
				\State $N++$
			\EndFor
			\State $\textbf{RemapInstanceID}(\mathcal{L}_{ID}, G)$
			\State \textbf{return} $\mathcal{L}_{ID}$
		\end{algorithmic}
	\end{algorithm}
	
	\subsection{Ellipsoidal Neighbor Searching}
	\label{sec:ellipsoidal-neighbor}
	After obtaining the semantic and known instance segmentation results, we conduct point cloud clustering on the part that belongs to neither the background, such as road, parking, and sidewalk, nor the known instance, such as pedestrians and vehicles.
	We propose ellipsoidal neighbor searching to accurately determine clusters in LiDAR scans.
	The ellipsoidal neighbor set $\mathcal{E}_\v{p}$ of point $\v{p}$ is shown in~\figref{fig:ellipsoidal-perspective}, where $\v{O}$ is the optical center of LiDAR.
	To generate this neighbor set, we first project point $\v{p}$ to the $X\v{O}Y$ plane getting $\v{p^{'}}$.
	The ellipsoidal axis $\alpha$ is then set as the same straight line as $\v{O}\v{p^{'}}$, and another ellipsoidal axis $\gamma$ is perpendicular to the $X\v{O}Y$ plane.
	The rest ellipsoidal axis $\beta$ is perpendicular to the plane composed of $\alpha$ and $\gamma$.
	Three ellipsoidal axes intersect at point $\v{p^{'}}$.
	Based on these three ellipsoidal axes, we can obtain their corresponding ellipsoid $\mathcal{E}_{\v{p^{'}}}$.
	$\mathcal{E}_{\v{p^{'}}}$ is then translated along the $Z$-axis until point $\v{p^{'}}$ and point $\v{p}$ coincide.
	The translated ellipsoid is the actual ellipsoidal neighbor $\mathcal{E}_\v{p}$ of point $\v{p}$.
	
	A point $\v{p_{i_n}} = (x, y, z)$ inside the ellipsoidal $\mathcal{E}_\v{p}$ is a neighboring point of $\v{p} = (x_p, y_p, z_p)$, satisfying the following formula:
	\begin{equation}
	\label{equ:equ1}
	\begin{split}
		\frac{((x - x_p)\cos(\lambda) + (y - y_p)\sin(\lambda)) ^ {2}}{a ^ {2}}\\
		+\frac{((x_p - x)\sin(\lambda) + (y - y_p)\cos(\lambda)) ^ {2}}{b ^ {2}}\\ +\frac{(z - z_p) ^ {2}}{c ^ {2}} \leq 1
	\end{split}
	\end{equation}
	where $\lambda$ is the angle between $\v{O}\v{p^{'}}$ and the positive $X$-axis.
	$a = \frac{\rho}{2}$, $b = \tan(\frac{\theta}{2}) * d$, and $c = \tan(\frac{\varphi}{2}) * d$ are half the lengths of axes $\alpha$, $\beta$, and $\gamma$, respectively.
	$d$ represents the length of segment $\v{O}\v{p^{'}}$.
	$\rho$, $\theta$, $\varphi$ are adjustable parameters used to control the scale of the ellipsoidal neighbor $\mathcal{E}_\v{p}$.

	Since the lengths of axes $\beta$ and $\gamma$ are proportional to the distance $d$ from LiDAR to each point in the $X\v{O}Y$ plane, the proposed ellipsoidal neighbor $\mathcal{E}_\v{p}$ is dynamically adjusted.
	Thus, the neighbor fits the characteristic of LiDAR scans, where the farther a point from LiDAR, the larger the distance from its neighbors in the horizontal and vertical directions.
	
	\begin{algorithm}[th]
		\caption{Diffuse Searching for Refinement}
		\label{alg:refinement}
		\begin{algorithmic}[1]
			\Require List of known instances ID $\mathcal{L}_{ID_k}$, point cloud of known instances $\mathcal{P}_k$, fixed searching radius $r$
			\Ensure List of refined known instances ID  $\mathcal{L}_{ID_{rk}}$
			\State \textbf{Initialization:} $\mathcal{L}_{ID_{rk}} \leftarrow \{0\}_{\mathcal{P}_k\text{.size()}}$, $\mathcal{L}_{NUM_E} \leftarrow$ list of existing ID number of known instances, Kd-tree $T_{\mathcal{P}_k}$ for $\mathcal{P}_k$, list of ID number of known instances needed to be refined $\mathcal{L}_{NUM_R}$, list of indexes of points needed to be queried $\mathcal{L}_{Q_q}$, list of indexes of points needed to be refined $\mathcal{L}_{Q_r}$
			\While {$\mathcal{L}_{NUM_E} \neq \emptyset$}
			\State $j_e = \mathcal{L}_{NUM_E}\text{.pop()}$
			\State $\mathcal{L}_{NUM_R}\text{.push}(j_e)$
			\State $\mathcal{L}_{Q_e} = \text{where}(\mathcal{L}_{ID_k} = j_e)$
			\State $\mathcal{L}_{Q_q}\text{.push}(\mathcal{L}_{Q_e})$
			\State $\mathcal{L}_{Q_r}\text{.push}(\mathcal{L}_{Q_e})$
			\While {$\mathcal{L}_{Q_q} \neq \emptyset$}
			\State $i_q = \mathcal{L}_{Q_q}\text{.pop()}$
			\State $\mathcal{L}_{Q_n} = T_{\mathcal{P}_k}.\textbf{FindNeighbor}(\mathcal{P}_k[i_q], r)$
			\State $\mathcal{L}_{NUM_N} \leftarrow \mathcal{L}_{ID_k}[\mathcal{L}_{Q_n}] \cap \mathcal{L}_{NUM_E}$
			\For  {each instance ID $j_n \in \mathcal{L}_{NUM_N}$}
			\If {$j_n \in \mathcal{L}_{NUM_R}$}
			\State \textbf{continue}
			\EndIf
			\State $\mathcal{L}_{Q_e} =$ where($\mathcal{L}_{ID_k} = j_n$)
			\State $\mathcal{L}_{Q_q}\text{.push}(\mathcal{L}_{Q_e})$
			\State $\mathcal{L}_{Q_r}\text{.push}(\mathcal{L}_{Q_e})$
			\State $\mathcal{L}_{NUM_E}\text{.delete}(j_n)$
			\EndFor
			\EndWhile
			\State $\mathcal{L}_{ID_{rk}}[\mathcal{L}_{Q_r}] = \textbf{EllipsoidalClustering}(\mathcal{P}_k[\mathcal{L}_{Q_r}])$
			\State $\mathcal{L}_{Q_r}\text{.clear}()$
			\State $\mathcal{L}_{NUM_R}\text{.clear}()$
			\EndWhile
			\State \textbf{return} $\mathcal{L}_{ID_{rk}}$
		\end{algorithmic}
	\end{algorithm}

	\subsection{Ellipsoidal Clustering}
	\label{sec:ellipsoidal-clustering}
	Based on the ellipsoidal neighbors, we then propose a clustering method to achieve accurate LiDAR instance clustering, as depicted in~\algref{alg:ellipsoidal-clustering}.
	For each query point $\v{p}_{i_q}$ in point cloud $\mathcal{P}$ to be clustered, we calculate the corresponding lengths $a$, $b$, and $c$ of the ellipsoidal semi-axes (line 11) as defined in \secref{sec:ellipsoidal-neighbor}.
	Then, we utilize Kd-tree to search points, which might belong to the ellipsoidal neighbor, within a certain radius of point $\v{p}_{i_q}$. To do so, we can avoid frequently querying the whole point cloud in the following steps and reduce the calculation cost (line 12).
	The searching radius is set to the maximum of $a$, $b$, and $c$, so that we do not miss any points that might be in the ellipsoidal neighbor, like the points inside the orange circle shown in~\figref{fig:ellipsoidal-bev}.
	For points in the searching radius, we utilize~\eqref{equ:equ1} to determine whether they are in the ellipsoidal neighbor $\mathcal{E}_{\v{p}_{i_q}}$ of point $\v{p}_{i_q}$ (lines 14-15).
	Those points within $\mathcal{E}_{\v{p}_{i_q}}$ are considered to be in the same cluster as point $\v{p}_{i_q}$, as the yellow and purple points inside the ellipse shown in~\figref{fig:ellipsoidal-bev}.
	In addition, these points are also added to the query list for further ellipsoidal neighbor searching (line 24), except that the distances between these points and point $\v{p}_{i_q}$ are less than the minimum of $a$, $b$, and $c$ (line 21).
	In this case, we regard these points as the middle part of the current cluster and do not need to be further extended (line 22), like purple points in~\figref{fig:ellipsoidal-bev}.
	By doing so, our ellipsoidal-based clustering is both efficient and accurate.

	To handle the situation when there are points with different instance IDs in the same ellipsoidal neighbor, we build a graph $G$ to merge these corresponding instances into the same one.
	We consider each instance ID as a vertex of the graph.
	For points with different instance IDs in the same ellipsoidal neighbor, we add edges between vertexes corresponding to these IDs (line 17).
	After all ellipsoidal neighbors of the query points are searched, the instance IDs are remapped by looking up the graph, where instances with connected instance IDs in the graph are merged into the same cluster (line 26).
	
	\begin{figure}[t]
		\centering
		\subfloat[]{
			\includegraphics[width=4.0cm]{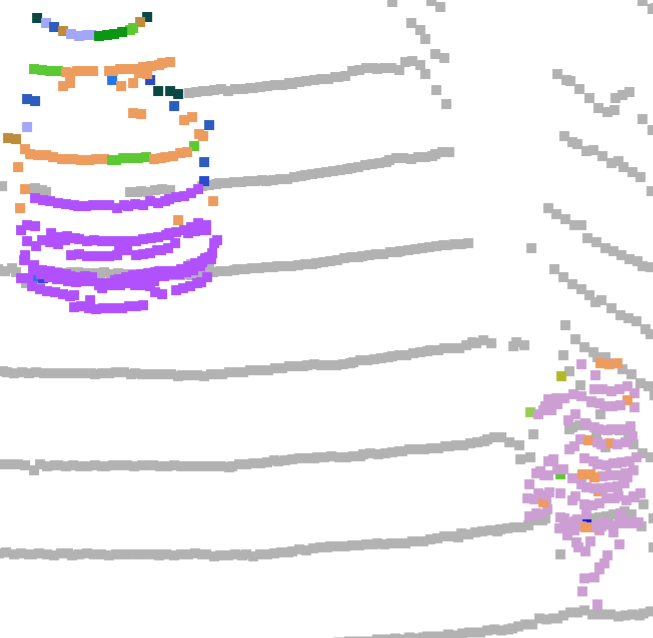}
			\label{fig:raw-known}
		}%
		\subfloat[]{
			\includegraphics[width=4.0cm]{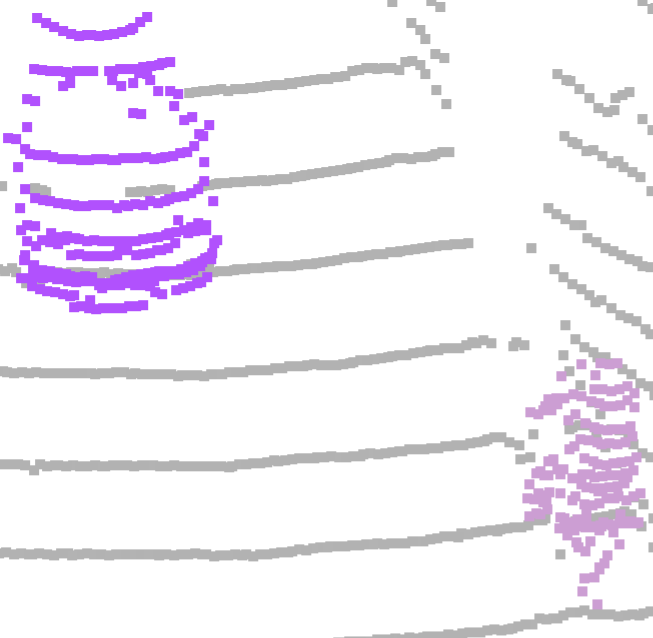}
			\label{fig:refined-known}
		}%
		\caption{The results of diffuse searching for known instances refinement. Each instance is marked with a unique color in one figure, and instances with the same colors in different figures are the same ones in the real world. (a) The raw known instances. The car on the top and the bicyclist below are over-segmented. (b) The refined known instances. The over-segmented parts are re-clustered into the correct instances.}
		\label{fig:diffuse-searching}
		\vspace{-0.2cm}
	\end{figure}
	
	\subsection{Diffuse Searching for Known Instances Refinement}
	\label{sec:refinement}
	To tackle the over-segmentation problem of the known instances from the close-set panoptic segmentation, we propose the diffuse searching method for refinement, as depicted in~\algref{alg:refinement}.
	Similar to the graph merging for unknown instances, the main idea of diffuse searching is also to find adjacent points with different instance IDs and diffuse the searching scope to points marked with those instance IDs.
	Specifically, for each cluster, we first utilize Kd-tree with a fixed searching radius to query all points within this cluster (line 5) and find the neighboring points (line 10).
	If there are neighboring points belonging to other clusters, we mark these clusters (lines 11-14) and keep searching for their neighboring points (lines 8-18).
	Until no point of new cluster is found, we then leverage our ellipsoidal clustering method to segment those searched points (line 19).
	We repeat the above steps till all original known instances are searched and re-clustered.
	~\figref{fig:diffuse-searching} shows the results of diffuse searching for refinement.

	\begin{table*}[t]
		\caption{$S_{assoc}$ of the open-world instance segmentation framework in the TEST set.}
		\label{tab:association-framework-test}
		\renewcommand\arraystretch{1.2}
		\setlength{\tabcolsep}{4pt}
		\small
		\begin{center}
			\begin{tabular}{l|cccccccc|ccc}
				\hline
				\textbf{Method} & car & bicycle & motorcycle & truck & other-vehicle & person & bicyclist & motorcyclist & known & unknown & all \\
				\hline
				Hu et al. & 0.755 & 0.200 & 0.539 & 0.512 & 0.509 & 0.503 & 0.693 & \textbf{0.793} & 0.697 & 0.587 & 0.678\\
				3DUIS & 0.762 & 0.203 & 0.552 & 0.722 & \textbf{0.737} & \textbf{0.684} & 0.788 & 0.773 & 0.720 & 0.599 & 0.699\\
				\hline
				\textbf{Ours} & \textbf{0.879} & \textbf{0.316} & \textbf{0.563} & \textbf{0.755} & 0.641 & 0.587 & \textbf{0.812} & 0.768 & \textbf{0.819} & \textbf{0.692} & \textbf{0.797}\\
				\hline
			\end{tabular}
		\end{center}
	\vspace{-0.2cm}
	\end{table*}

	\section{Experimental Results}
	The following experiments support our three claims that (i)~our method achieves the state-of-the-art LiDAR open-world instance segmentation (OIS) performance; (ii) our proposed ellipsoidal clustering method outperforms baseline methods in segmenting instances on LiDAR data, and (iii)~our framework is highly modular and flexible with different panoptic and instance segmentation methods.
	
	\subsection{Experiment Setups}
	We evaluate our framework on the SemanticKITTI open-world LiDAR instance segmentation dataset~\cite{nunes2022ral}.
	This dataset is the extension of SemanticKITTI~\cite{behley2019iccv}, with additionally labeled unknown instances in the validation set and test set.
	In this dataset, the evaluation of unknown instances is only accessible on the hidden server side, increasing the challenge level.
	We follow the benchmark~\cite{nunes2022ral} using intersection over union (IoU), Recall, and the association metric $S_{assoc}$ to measure the quantitative results.
	The association metric considers true positive associations and IoU, and the higher the $S_{assoc}$ scores, the better a method for performing OIS.
	
	Before clustering, we use the predicted semantic labels from semantic segmentation to remove background points belonging to \textit{road}, \textit{parking}, \textit{sidewalk}, \textit{other-ground}, \textit{terrain}, and \textit{vegetation}, which are typical background classes of SemanticKITTI~\cite{behley2019iccv}. We set our ellipsoidal clustering parameters to $\rho = 2.0$, $\theta = 2.0^\circ$, and $\varphi = 7.5^\circ$ in all the following experiments.

	\begin{table*}[t]
		\caption{The IoU and Recall of the open-world instance segmentation frameworks in the TEST set.}
		\label{tab:iou-recall-test}
		\renewcommand\arraystretch{1.2}
		\setlength{\tabcolsep}{3pt}
		\small
		\begin{center}
			\begin{tabular}{l|c|c|c|c|c|c|c|c|c|c|c|c}
				\hline
				\multirow{3}*{\textbf{Method}} & \multicolumn{6}{c|}{IoU} & \multicolumn{6}{c}{Recall}\\
				\cline{2-13} & \multicolumn{2}{c|}{90\%} & \multicolumn{2}{c|}{70\%} & \multicolumn{2}{c|}{50\%} & \multicolumn{2}{c|}{90\%} & \multicolumn{2}{c|}{70\%} & \multicolumn{2}{c}{50\%} \\
				\cline{2-13} & known & unknown & known & unknown & known & unknown & known & unknown & known & unknown & known & unknown \\
				\hline
				Hu et al. & 0.510 & 0.410 & 0.670 & 0.562 & 0.721 & 0.596 & 0.528 & 0.424 & 0.724 & 0.608 & 0.806 & 0.664\\
				3DUIS & 0.461 & 0.381 & 0.686 & 0.546 & 0.745 & 0.591 & 0.481 & 0.393 & 0.755 & 0.595 & 0.850 & 0.668\\
				\hline
				\textbf{Ours} & \textbf{0.710} & \textbf{0.494} & \textbf{0.797} & \textbf{0.667} & \textbf{0.824} & \textbf{0.693} & \textbf{0.730} & \textbf{0.512} & \textbf{0.835} & \textbf{0.719} & \textbf{0.880} & \textbf{0.761}\\
				\hline
			\end{tabular}
		\end{center}
	\vspace{-0.2cm}
	\end{table*}

	\begin{table}[t]
		\caption{$S_{assoc}$ scores of different clustering methods for unknown instances in the VALIDATION set.}
		\label{tab:ablation-with-known-clustering-val}
		\renewcommand\arraystretch{1.2}
		\setlength{\tabcolsep}{8pt}
		\small
		\begin{center}
			\begin{tabular}{l|ccc}
				\hline
				\textbf{Method} & known & unknown & all \\
				\hline
				EC & \textbf{0.846} & 0.769 & 0.830\\
				CVC & \textbf{0.846} & 0.782 & 0.833\\
				HDBSCAN & \textbf{0.846} & 0.789 & 0.834\\
				\hline
				\textbf{ElC (ours)} & \textbf{0.846} & \textbf{0.793} & \textbf{0.835}\\
				\hline
			\end{tabular}
		\end{center}
	\end{table}
	
	\subsection{Open-World Instance Segmentation Benchmark}	
	In the first experiment, we show that our method outperforms the state-of-the-art LiDAR OIS methods on the SemanticKITTI LiDAR OIS benchmark.
	Our framework using the DS-Net as the close-set panoptic segmentation method, our ellipsoidal clustering as the clustering method, and the diffuse searching method for instance refinement, is denoted as~\textit{ours}.
	We compare our method with the state-of-the-art LiDAR OIS methods: 3DUIS~\cite{nunes2022ral} and the method by Hu~\etalcite{hu2020ral}.
	3DUIS leverages HDBSCAN~\cite{campello2013pakdd} to generate instance proposals, then segments the background using the point-wise features and the graph representation.
	On the other hand, Hu's method uses the pre-processed LiDAR points with no background to generate a mass of instance proposals, and it selects the instances with the highest evaluated objectness scores as the final ones.

	The quantitative results in the test set of the benchmark are shown in \tabref{tab:association-framework-test}.	
	As can be seen, our method significantly outperforms the state-of-the-art by around 10\% in known, unknown, and overall instance segmentation in terms of association scores.
	
	In line with~\cite{nunes2022ral}, we also provide the IoU and Recall results filtered by different IoU thresholds, 90\%, 70\%, and 50\%, of our method with those of 3DUIS and Hu~et al. on the test set of the benchmark as shown in~\tabref{tab:iou-recall-test}. 
	As can be seen, our method surpasses the other two methods in all items under different thresholds. It indicates that instances segmented by our method match the actual objects more accurately and cover most of the points belonging to those objects.

	\begin{table*}[t]
		\setlength{\abovecaptionskip}{0.0cm}
		\setlength{\belowcaptionskip}{0.0cm}
		\caption{$S_{assoc}$ with different clustering methods for all foreground points in the VALIDATION set.}
		\label{tab:ablation-foreground-clustering-val}
		\renewcommand\arraystretch{1.2}
		\setlength{\tabcolsep}{3pt}
		\small
		\begin{center}
			\begin{tabular}{l|cccccccc|ccc}
				\hline
				\textbf{Method} & car & bicycle & motorcycle & truck & other-vehicle & person & bicyclist & motorcyclist & known & unknown & all \\
				\hline
				HDBSCAN & 0.692 & \textbf{0.343} & 0.680 & 0.573 & 0.542 & \textbf{0.739} & 0.875 & 0.618 & 0.676 & \textbf{0.789} & 0.699\\
				EC & 0.860 & 0.260 & 0.613 & 0.750 & 0.659 & 0.492 & 0.840 & 0.531 & 0.801 & 0.746 & 0.790\\
				CVC & 0.853 & 0.279 & 0.627 & 0.718 & 0.656 & 0.556 & 0.896 & 0.621 & 0.801 & 0.772 & 0.795\\
				\hline
				\textbf{ElC (ours)} & \textbf{0.887} & 0.314 & \textbf{0.734} & \textbf{0.753} & \textbf{0.678} & 0.594 & \textbf{0.904} & \textbf{0.668} & \textbf{0.835} & \textbf{0.789} & \textbf{0.826}\\
				\hline
			\end{tabular}
		\end{center}
	\vspace{-0.4cm}
	\end{table*}
	
	\subsection{Study on Different Clustering Methods}
	To further validate the effectiveness of each module in our framework, we conduct multiple studies on the validation set of the benchmark (due to the 10 times submission limit of the hidden test set).
	The first experiment studies the impact of different clustering methods and shows that our proposed ellipsoidal clustering (ElC) fits the OIS task best. It also shows that our framework can work with different clustering methods.
	We compare our method with three other clustering methods: Euclidean clustering (EC)~\cite{rusu2011icra}, HDBSCAN~\cite{campello2013pakdd}, and curved-voxel clustering (CVC)~\cite{park2019iros}.
	Euclidean clustering searches neighboring points with a fixed radius, and CVC considers the neighboring points in 3$^3$ curved voxels near the target voxel to be in the same cluster.
	HDBSCAN leverages the density of point clouds to  implement clustering.

	For a fair comparison, we set the searching ranges of every cluster method as similar to each other as possible.
	Thus, the searching radius of Euclidean clustering is set as half of the fixed ellipsoidal axis $\rho$, which is 1.0 in this case.
	As for CVC, the parameters $\Delta\rho$, $\Delta\theta$, and $\Delta\varphi$, which respectively correspond to the heading direction, horizontal direction, and vertical direction, are set as one-third of the ellipsoidal parameters $\rho$, $\theta$, and $\varphi$.
	HDBSCAN does not use any distance-related parameters.
	We use DS-Net as the close-set panoptic segmentation method for all clustering methods and do not apply the diffuse searching method for refinement.

	\tabref{tab:ablation-with-known-clustering-val} shows the segmentation results for unknown instances.
	Since we keep the same known instances for all methods, the scores of the known instances are the same, while our proposed ellipsoidal clustering method outperforms other clustering methods in segmenting the unknown instances and the overall instances.
	
\begin{figure*}[t]
		\centering
		\subfloat[HDBSCAN]{
			\includegraphics[width=3.3cm]{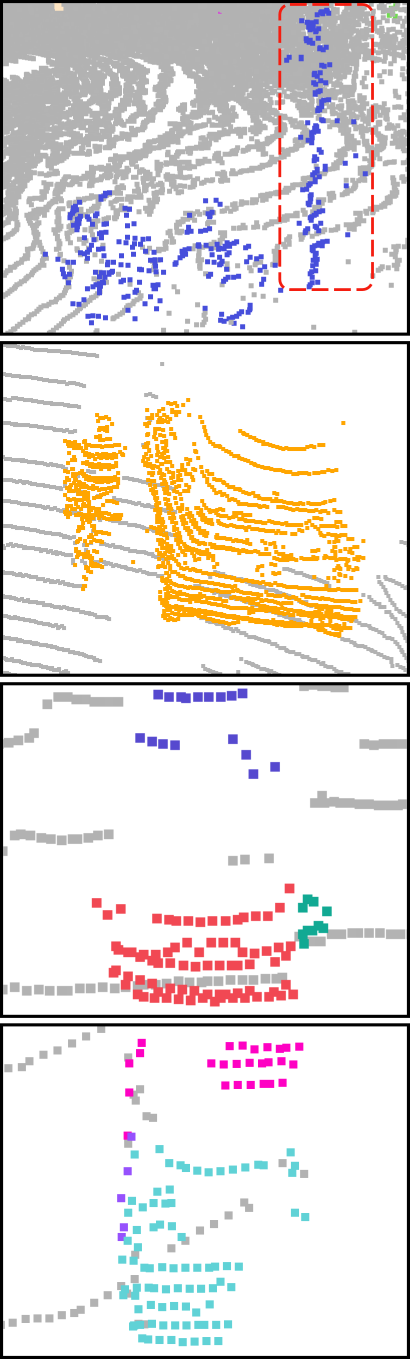}
			\label{fig:clustering-compare-hdbscan}
		}%
		\subfloat[EC]{
			\includegraphics[width=3.3cm]{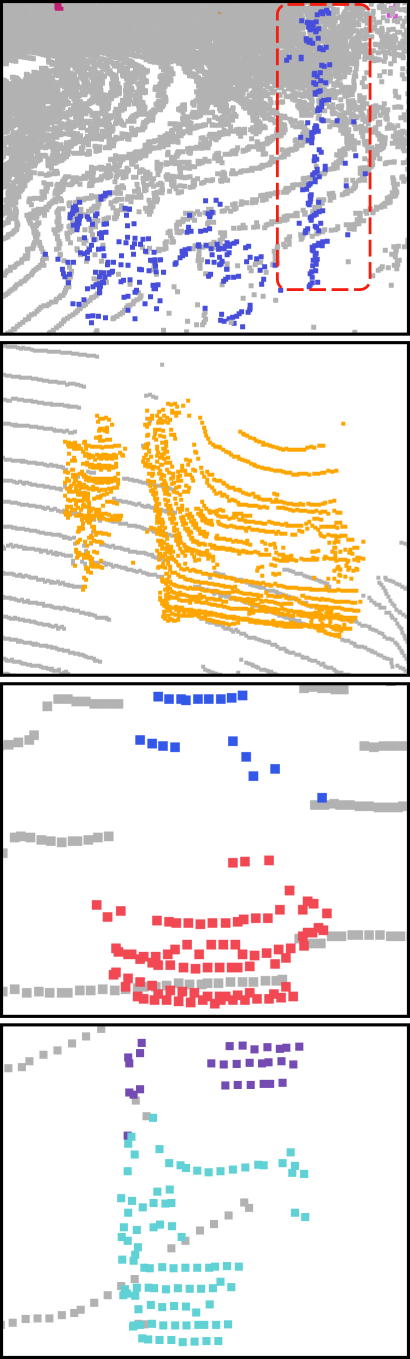}
			\label{fig:clustering-compare-tec}
		}%
		\subfloat[CVC]{
			\includegraphics[width=3.3cm]{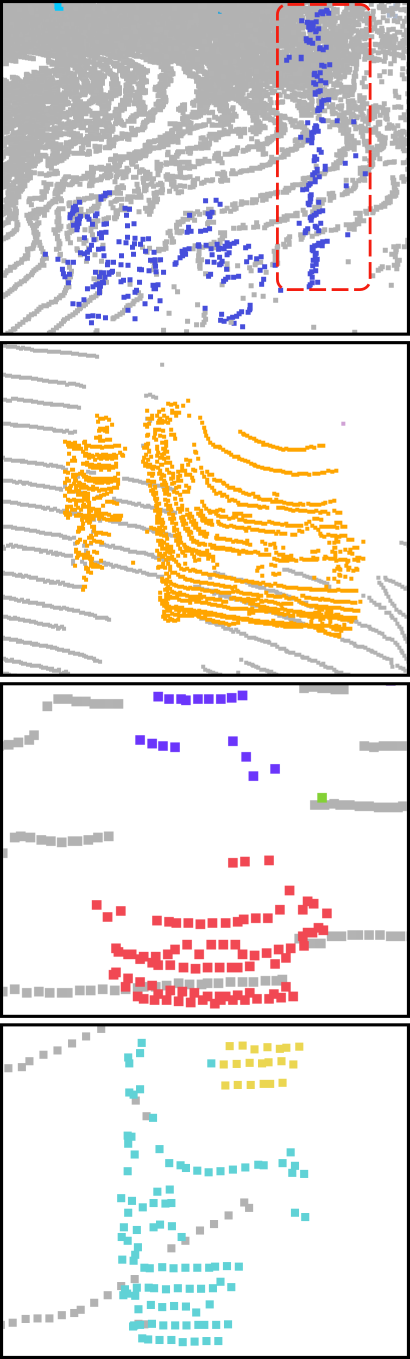}
			\label{fig:clustering-compare-cvc}
		}%
		\subfloat[ElC (ours)]{
			\includegraphics[width=3.3cm]{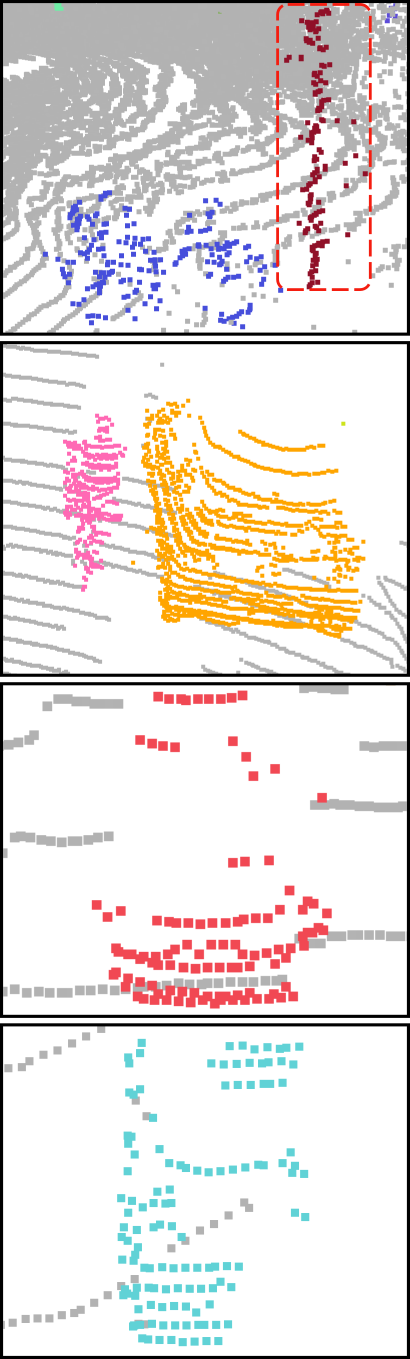}
			\label{fig:clustering-compare-elc}
		}%
		\subfloat[Ground truth]{
			\includegraphics[width=3.3cm]{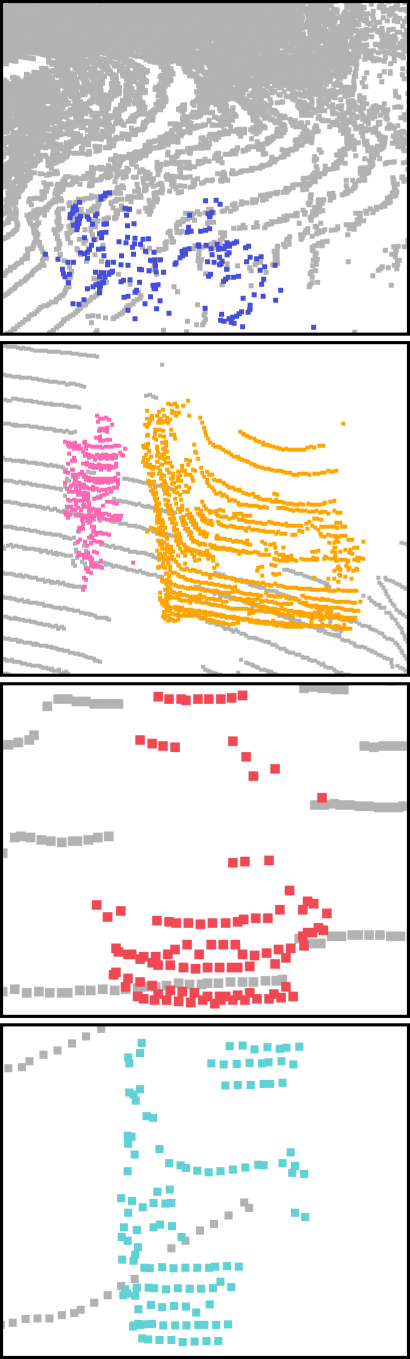}
			\label{fig:clustering-compare-gt}
		}%
		\caption{Visualization of the clustering results of foreground points from different clustering methods and the ground truth. Instances close to LiDAR, such as the bicycle and the nearby pole in the first row, the bicyclist and the car in the second row, are correctly segmented by our method, while under-segmented by other baseline methods. The last two rows show the segmentation results on the faraway cars. Our method consistently performs well because the proposed ellipsoidal clustering is adapted dynamically and accurately, while other methods over-segment the objects. The ground truth is from the close-set SemanticKITTI panoptic segmentation benchmark~\cite{behley2021icra}, which only annotates known instances, so the unknown instance in the first row (circled with red lines) is not accessible in the ground truth.}
		\label{fig:clustering-compare}
		\vspace{-0.1cm}
	\end{figure*}

	Additionally, to show the clustering capacity of our method in varying instances, we apply clustering methods to all foreground points for instance segmentation.
	The $S_{assoc}$ results are shown in \tabref{tab:ablation-foreground-clustering-val}.
	Our approach gets the highest $S_{assoc}$ in known instances, unknown instances, the overall instances, and most of the known classes in the validation set.
	These results validate that the segmenting effectiveness of our ellipsoidal clustering method is superior to that of other methods.
	\figref{fig:clustering-compare} visualizes the segmentation results from our method and the other three clustering methods.
	Each instance is marked with a unique color.
	Our method correctly segments the point cloud in both near and far distances from LiDAR, while other methods perform under-segmentation to the near point cloud and over-segmentation to the far point cloud.
	
	\begin{figure*}[th]
		\centering
		\includegraphics[width=0.9\textwidth]{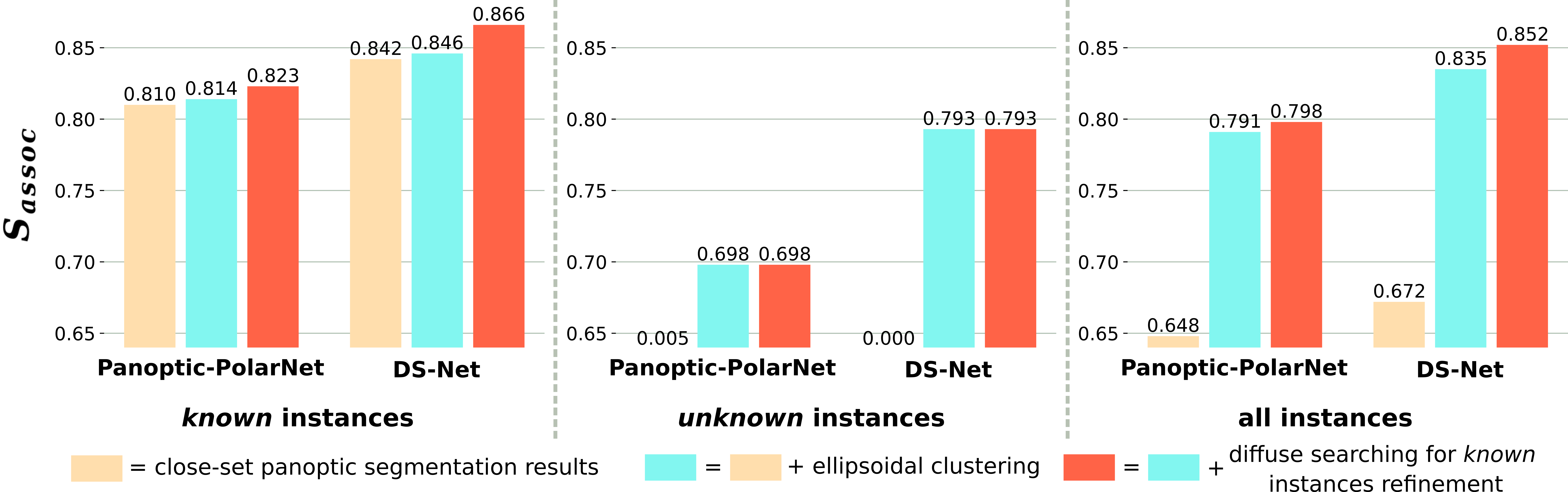}
		\caption{Ablation study of our framework. We sequentially add the ellipsoidal clustering method and the diffuse-searching instances refinement method to two different close-set panoptic segmentation (Panoptic-PolarNet and DS-Net). It shows the effectiveness of the proposed ellipsoidal clustering and the diffuse-searching refinement for improving the clustering performance on both unknown and known instance segmentation. Moreover, it supports that our framework can flexibly adjust the source of close-set panoptic segmentation results.}
		\label{fig:ablation-histogram}
		\vspace{-0.3cm}
	\end{figure*}

	\subsection{Ablation Study}
	\label{sec:as}
	In this section, we evaluate the influence of three parts of our framework on the overall framework: a) the changes of close-set panoptic segmentation methods, b) with/without the ellipsoidal clustering method, and c) with/without the diffuse searching method.
	To this end, we test two close-set panoptic segmentation methods, Panoptic-PolarNet~\cite{zhou2021cvpr} and DS-Net~\cite{hong2021cvpr}.
	Particularly, these two methods get 54.1 and 55.9 PQ scores in the close-set SemanticKITTI: Panoptic Segmentation dataset~\cite{behley2021icra}.
	We sequentially add the ellipsoidal clustering method and the diffuse-searching instances refinement method to both close-set panoptic segmentation methods.
	
	The ablation study results are depicted in \figref{fig:ablation-histogram}.
	As can be seen, our method works flexibly and effectively with different panoptic segmentation methods.
	Furthermore, our proposed ellipsoidal clustering method successfully handles the unknown instances and effectively improves the overall $S_{assoc}$ scores.
	After adding our ellipsoidal clustering method, the framework with DS-Net gets higher scores in the unknown instances, known instances, and overall instances.
	Meanwhile, the framework with Panoptic-PolarNet also gets competitive scores.
	In the third step, we add the refinement method, which further improves the results of known instances segmentation for both Panoptic-PolarNet and DS-Net.
	
	The overall results validate that our framework is highly modular and flexible with different panoptic segmentation methods. Besides, our framework performs better when more accurate semantic information and known instance information are provided. The ellipsoidal clustering method and the diffuse searching refinement method are proved to be effective for the unknown instance segmentation and the known instance segmentation.

	\section{CONCLUSION}
	In this paper, we proposed a flexible and effective open-world LiDAR instance segmentation framework, together with the novel ellipsoidal clustering and diffuse searching method, to accurately segment instances in the open world.
	This framework utilizes the close-set panoptic segmentation results to remove the background points in the raw LiDAR scan and generate known instances.
	The left unknown foreground points are further clustered by our proposed ellipsoidal clustering method, which considers the characteristic of LiDAR scans and dynamically adjusts the searching scope.
	Meanwhile, the diffuse searching method is proposed to refine the over-segmented instances.
	Our experimental results on the SemanticKITTI LiDAR OIS benchmark validated that our framework achieves the state-of-the-art LiDAR OIS performance. Multiple ablation studies showed that our framework is highly modular and flexible to different panoptic segmentation and clustering methods.
	
%	\addtolength{\textheight}{-12cm}   % This command serves to balance the column lengths

	%%%%%%%%%%%%%%%%%%%%%%%%%%%%%%%%%%%%%%%%%%%%%%%%%%%%%%%%%%%%%%%%%%%%%%%%%%%%%%%%

	%%%%%%%%%%%%%%%%%%%%%%%%%%%%%%%%%%%%%%%%%%%%%%%%%%%%%%%%%%%%%%%%%%%%%%%%%%%%%%%%

	%%%%%%%%%%%%%%%%%%%%%%%%%%%%%%%%%%%%%%%%%%%%%%%%%%%%%%%%%%%%%%%%%%%%%%%%%%%%%%%%

	%%%%%%%%%%%%%%%%%%%%%%%%%%%%%%%%%%%%%%%%%%%%%%%%%%%%%%%%%%%%%%%%%%%%%%%%%%%%%%%%

	\bibliographystyle{ieeetr}
	\bibliography{ms}

\begin{thebibliography}{10}

\bibitem{milioto2019iros}
A.~Milioto, I.~Vizzo, J.~Behley, and C.~Stachniss, ``{RangeNet++: Fast and
  Accurate LiDAR Semantic Segmentation},'' in {\em Proc. of the IEEE/RSJ Int.
  Conf. on Intelligent Robots and Systems (IROS)}, 2019.

\bibitem{li2022ral}
S.~Li, X.~Chen, Y.~Liu, D.~Dai, C.~Stachniss, and J.~Gall, ``{Multi-scale
  Interaction for Real-time LiDAR Data Segmentation on an Embedded Platform},''
  {\em IEEE Robotics and Automation Letters (RA-L)}, vol.~7, no.~2,
  pp.~738--745, 2022.

\bibitem{hong2021cvpr}
F.~Hong, H.~Zhou, X.~Zhu, H.~Li, and Z.~Liu, ``Lidar-based panoptic
  segmentation via dynamic shifting network,'' in {\em Proc.~of the IEEE/CVF
  Conf.~on Computer Vision and Pattern Recognition (CVPR)}, 2021.

\bibitem{sirohi2022tro}
K.~Sirohi, R.~Mohan, D.~Büscher, W.~Burgard, and A.~Valada, ``Efficientlps:
  Efficient lidar panoptic segmentation,'' {\em IEEE Trans.~on Robotics (TRO)},
  vol.~38, no.~3, pp.~1894--1914, 2022.

\bibitem{hu2020ral}
P.~Hu, D.~Held, and D.~Ramanan, ``Learning to optimally segment point clouds,''
  {\em IEEE Robotics and Automation Letters (RA-L)}, vol.~5, no.~2,
  pp.~875--882, 2020.

\bibitem{nunes2022ral}
L.~Nunes, X.~Chen, R.~Marcuzzi, A.~Osep, L.~Leal-Taixé, C.~Stachniss, and
  J.~Behley, ``Unsupervised class-agnostic instance segmentation of 3d lidar
  data for autonomous vehicles,'' {\em IEEE Robotics and Automation Letters
  (RA-L)}, vol.~7, no.~4, pp.~8713--8720, 2022.

\bibitem{wang2022cvpr}
W.~Wang, M.~Feiszli, H.~Wang, J.~Malik, and D.~Tran, ``Open-world instance
  segmentation: Exploiting pseudo ground truth from learned pairwise
  affinity,'' in {\em Proc.~of the IEEE/CVF Conf.~on Computer Vision and
  Pattern Recognition (CVPR)}, 2022.

\bibitem{hwang2021cvpr}
J.~Hwang, S.~W. Oh, J.-Y. Lee, and B.~Han, ``Exemplar-based open-set panoptic
  segmentation network,'' in {\em Proc.~of the IEEE/CVF Conf.~on Computer
  Vision and Pattern Recognition (CVPR)}, 2021.

\bibitem{wang2021cvpr}
X.~Wang, J.~Feng, B.~Hu, Q.~Ding, L.~Ran, X.~Chen, and W.~Liu,
  ``Weakly-supervised instance segmentation via class-agnostic learning with
  salient images,'' in {\em Proc.~of the IEEE/CVF Conf.~on Computer Vision and
  Pattern Recognition (CVPR)}, 2021.

\bibitem{wang2021iccv}
W.~Wang, M.~Feiszli, H.~Wang, and D.~Tran, ``Unidentified video objects: A
  benchmark for dense, open-world segmentation,'' in {\em Proc.~of the IEEE/CVF
  Intl.~Conf.~on Computer Vision (ICCV)}, 2021.

\bibitem{zhang2020icra}
F.~Zhang, C.~Guan, J.~Fang, S.~Bai, R.~Yang, P.~H. Torr, and V.~Prisacariu,
  ``Instance segmentation of lidar point clouds,'' in {\em Proc.~of the IEEE
  Intl.~Conf.~on Robotics \& Automation (ICRA)}, 2020.

\bibitem{chen2022tiv}
T.-H. Chen and T.~S. Chang, ``Rangeseg: Range-aware real time segmentation of
  3d lidar point clouds,'' {\em IEEE Trans.~on Intelligent Vehicles}, vol.~7,
  no.~1, pp.~93--101, 2022.

\bibitem{marcuzzi2023ral}
R.~Marcuzzi, L.~Nunes, L.~Wiesmann, J.~Behley, and C.~Stachniss, ``Mask-based
  panoptic lidar segmentation for autonomous driving,'' {\em IEEE Robotics and
  Automation Letters (RA-L)}, vol.~8, no.~2, pp.~1141--1148, 2023.

\bibitem{aygun2021}
M.~Aygün, A.~Ošep, M.~Weber, M.~Maximov, C.~Stachniss, J.~Behley, and
  L.~Leal-Taixé, ``4d panoptic lidar segmentation,'' in {\em Proc.~of the
  IEEE/CVF Conf.~on Computer Vision and Pattern Recognition (CVPR)}, 2021.

\bibitem{rusu2011icra}
R.~B. Rusu and S.~Cousins, ``3d is here: Point cloud library (pcl),'' in {\em
  Proc.~of the IEEE Intl.~Conf.~on Robotics \& Automation (ICRA)}, 2011.

\bibitem{klasing2008icra}
K.~Klasing, D.~Wollherr, and M.~Buss, ``A clustering method for efficient
  segmentation of 3d laser data,'' in {\em Proc.~of the IEEE Intl.~Conf.~on
  Robotics \& Automation (ICRA)}, 2008.

\bibitem{douillard2011icra}
B.~Douillard, J.~Underwood, N.~Kuntz, V.~Vlaskine, A.~Quadros, P.~Morton, and
  A.~Frenkel, ``On the segmentation of 3d lidar point clouds,'' in {\em
  Proc.~of the IEEE Intl.~Conf.~on Robotics \& Automation (ICRA)}, 2011.

\bibitem{bogoslavskyi2016iros}
I.~Bogoslavskyi and C.~Stachniss, ``Fast range image-based segmentation of
  sparse 3d laser scans for online operation,'' in {\em Proc.~of the IEEE/RSJ
  Intl.~Conf.~on Intelligent Robots and Systems (IROS)}, 2016.

\bibitem{zermas2017icra}
D.~Zermas, I.~Izzat, and N.~Papanikolopoulos, ``Fast segmentation of 3d point
  clouds: A paradigm on lidar data for autonomous vehicle applications,'' in
  {\em Proc.~of the IEEE Intl.~Conf.~on Robotics \& Automation (ICRA)}, 2017.

\bibitem{chen2019iros}
X.~Chen, A.~Milioto, E.~Palazzolo, P.~Giguère, J.~Behley, and C.~Stachniss,
  ``{SuMa++: Efficient LiDAR-based Semantic SLAM},'' in {\em Proc. of the
  IEEE/RSJ Int. Conf. on Intelligent Robots and Systems (IROS)}, 2019.

\bibitem{paigwar2020iros}
A.~Paigwar, O.~Erkent, D.~Sierra-Gonzalez, and C.~Laugier, ``Gndnet: Fast
  ground plane estimation and point cloud segmentation for autonomous
  vehicles,'' in {\em Proc.~of the IEEE/RSJ Intl.~Conf.~on Intelligent Robots
  and Systems (IROS)}, 2020.

\bibitem{behley2013iros}
J.~Behley, V.~Steinhage, and A.~B. Cremers, ``Laser-based segment
  classification using a mixture of bag-of-words,'' in {\em Proc.~of the
  IEEE/RSJ Intl.~Conf.~on Intelligent Robots and Systems (IROS)}, 2013.

\bibitem{che2018jprs}
E.~Che and M.~J. Olsen, ``Multi-scan segmentation of terrestrial laser scanning
  data based on normal variation analysis,'' {\em ISPRS Journal of
  Photogrammetry and Remote Sensing (JPRS)}, vol.~143, pp.~233--248, 2018.

\bibitem{huang2019rs}
M.~Huang, P.~Wei, and X.~Liu, ``An efficient encoding voxel-based segmentation
  (evbs) algorithm based on fast adjacent voxel search for point cloud plane
  segmentation,'' {\em Remote Sensing}, vol.~11, no.~23, 2019.

\bibitem{klasing2009icra}
K.~Klasing, D.~Althoff, D.~Wollherr, and M.~Buss, ``Comparison of surface
  normal estimation methods for range sensing applications,'' in {\em Proc.~of
  the IEEE Intl.~Conf.~on Robotics \& Automation (ICRA)}, 2009.

\bibitem{jagannathan2007pami}
A.~Jagannathan and E.~L. Miller, ``Three-dimensional surface mesh segmentation
  using curvedness-based region growing approach,'' {\em IEEE Transactions on
  Pattern Analysis and Machine Intelligence (PAMI)}, vol.~29, no.~12,
  pp.~2195--2204, 2007.

\bibitem{ester1996kdd}
M.~Ester, H.~Kriegel, J.~Sander, and X.Xu, ``A density-based algorithm for
  discovering clusters in large spatial databases with noise.,'' in {\em Proc.
  of the Conf. on Knowledge Discovery and Data Mining (KDD)}, 1996.

\bibitem{campello2013pakdd}
R.~J. G.~B. Campello, D.~Moulavi, and J.~Sander, ``Density-based clustering
  based on hierarchical density estimates,'' in {\em Advances in Knowledge
  Discovery and Data Mining (PAKDD)}, 2013.

\bibitem{park2019iros}
S.~Park, S.~Wang, H.~Lim, and U.~Kang, ``Curved-voxel clustering for accurate
  segmentation of 3d lidar point clouds with real-time performance,'' in {\em
  Proc.~of the IEEE/RSJ Intl.~Conf.~on Intelligent Robots and Systems (IROS)},
  2019.

\bibitem{nunes2022seg}
L.~Nunes, R.~Marcuzzi, X.~Chen, J.~Behley, and C.~Stachniss, ``{SegContrast: 3D
  Point Cloud Feature Representation Learning through Self-supervised Segment
  Discrimination},'' {\em IEEE Robotics and Automation Letters (RA-L)}, vol.~7,
  no.~2, pp.~2116--2123, 2022.

\bibitem{chen2021mos}
X.~Chen, S.~Li, B.~Mersch, L.~Wiesmann, J.~Gall, J.~Behley, and C.~Stachniss,
  ``{Moving Object Segmentation in 3D LiDAR Data: A Learning-based Approach
  Exploiting Sequential Data},'' {\em IEEE Robotics and Automation Letters
  (RA-L)}, vol.~6, pp.~6529--6536, 2021.

\bibitem{jin2022iros}
S.~Jin, Z.~Wu, C.~Zhao, J.~Zhang, G.~Peng, and D.~Wang, ``Sectionkey: 3-d
  semantic point cloud descriptor for place recognition,'' in {\em Proc.~of the
  IEEE/RSJ Intl.~Conf.~on Intelligent Robots and Systems (IROS)}, 2022.

\bibitem{wong2020corl}
K.~Wong, S.~Wang, M.~Ren, M.~Liang, and R.~Urtasun, ``Identifying unknown
  instances for autonomous driving,'' in {\em Proc.~of the Conf.~on Robot
  Learning (CoRL)}, 2020.

\bibitem{qi2017nips}
C.~R. Qi, L.~Yi, H.~Su, and L.~J. Guibas, ``Pointnet++: Deep hierarchical
  feature learning on point sets in a metric space,'' in {\em Proc.~of the
  Advances in Neural Information Processing Systems (NIPS)}, vol.~30, 2017.

\bibitem{chen2021ral}
J.~Chen, Z.~Kira, and Y.~K. Cho, ``Lrgnet: Learnable region growing for
  class-agnostic point cloud segmentation,'' {\em IEEE Robotics and Automation
  Letters (RA-L)}, vol.~6, no.~2, pp.~2799--2806, 2021.

\bibitem{zhou2021cvpr}
Z.~Zhou, Y.~Zhang, and H.~Foroosh, ``Panoptic-polarnet: Proposal-free lidar
  point cloud panoptic segmentation,'' in {\em Proc.~of the IEEE/CVF Conf.~on
  Computer Vision and Pattern Recognition (CVPR)}, 2021.

\bibitem{behley2019iccv}
J.~Behley, M.~Garbade, A.~Milioto, J.~Quenzel, S.~Behnke, C.~Stachniss, and
  J.~Gall, ``Semantickitti: A dataset for semantic scene understanding of lidar
  sequences,'' in {\em Proc.~of the IEEE/CVF Intl.~Conf.~on Computer Vision
  (ICCV)}, 2019.

\bibitem{behley2021icra}
J.~Behley, A.~Milioto, and C.~Stachniss, ``A benchmark for lidar-based panoptic
  segmentation based on kitti,'' in {\em Proc.~of the IEEE Intl.~Conf.~on
  Robotics \& Automation (ICRA)}, 2021.

\end{thebibliography}

\end{document}